\documentclass[final]{cvpr}
\makeatletter
\@namedef{ver@everyshi.sty}{}
\makeatother
\usepackage{times}
\usepackage{epsfig}
\usepackage{graphicx}
\usepackage{amsmath}
\usepackage{amssymb}
\usepackage{subcaption}
\usepackage{enumitem}

\usepackage{etoolbox}
\usepackage{silence}
\makeatletter
\robustify\@latex@warning@no@line
\makeatother
\usepackage{authblk}

\usepackage{svg}

\usepackage[pagebackref=true,breaklinks=true,colorlinks,bookmarks=false]{hyperref}

\def\cvprPaperID{28} 
\def\confYear{CVPR 2021}

\begin{document}

\title{Vision-based Behavioral Recognition of Novelty Preference in Pigs}

\author[1]{Aniket Shirke}
\author[2]{Rebecca Golden}
\author[1]{Mrinal Gautam}
\author[3]{Angela Green-Miller}
\author[1]{Matthew Caesar}
\author[2]{Ryan N. Dilger}

\affil[1]{Department of Computer Science, University of Illinois at Urbana-Champaign}
\affil[2]{Department of Animal Sciences, University of Illinois at Urbana-Champaign}
\affil[3]{Department of Agricultural \& Biological Engineering, University of Illinois at Urbana-Champaign}

\affil[ ]{\small \texttt {\{anikets, rkgolde2, mgautam2, angelag, caesar, rdilger2\}@illinois.edu}}

\maketitle

\begin{abstract}
   Behavioral scoring of research data is crucial for extracting domain-specific metrics but is bottlenecked on the ability to analyze enormous volumes of information using human labor. Deep learning is widely viewed as a key advancement to relieve this bottleneck. We identify one such domain, where deep learning can be leveraged to alleviate the process of manual scoring. Novelty preference paradigms have been widely used to study recognition memory in pigs, but analysis of these videos requires human intervention. We introduce a subset of such videos in the form of the `Pig Novelty Preference Behavior' (PNPB) dataset that is fully annotated with pig actions and keypoints. In order to demonstrate the application of state-of-the-art action recognition models on this dataset, we compare LRCN, C3D, and TSM on the basis of various analytical metrics and discuss common pitfalls of the models. Our methods achieve an accuracy of 93\% and a mean Average Precision of 96\% in estimating piglet behavior.  
   We open-source our code and annotated dataset at \url{https://github.com/AIFARMS/NOR-behavior-recognition}
\end{abstract}

\section{Introduction}

In recent years, video-based action recognition has become a popular field of research in deep learning. Over the decade, this technology has evolved to produce reliable outcomes in real-time activity recognition based on a large collection of data sets such as UCF101 \cite{soomro2012ucf101} and Sports1M \cite{KarpathyCVPR14}. There are a large number of applications in the domain of robotics, healthcare, human-computer interaction systems, traffic monitoring, and control, etc \cite{action-survey}. 

One primary application of deep learning is to automate the annotation procedure for datasets. Visual research data, consisting of videos, often need to be annotated manually for extracting insightful metrics. Instead of manually annotating videos by utilizing human labor, these tasks can be offloaded to a robust deep learning model that can score videos in a much more efficient manner. Such deep learning-based behavioral analysis has reached human accuracy and outperformed commercial solutions \cite{annotatedeep}.

The use of behavioral tasks in animal research, such as novel object recognition (NOR) in pigs, permits functional assessment of working memory in animal subjects. Importantly, NOR has proven to be sensitive to early-life dietary patterns, thereby confirming that such environmental influences can alter brain development \cite{fleming2018dietary}\cite{fleming2019dietary}. Here, we present a dataset and approach that can help to develop deep learning models for automatically annotating NOR test videos. We also investigate the effects of state-of-the-art action recognition models on NOR metrics and present findings where technical challenges remain.

\section{Novel Object Recognition Task for Pigs}
\label{sec:nor}

\begin{figure}[h]
\begin{center}
   \includegraphics[width=\linewidth]{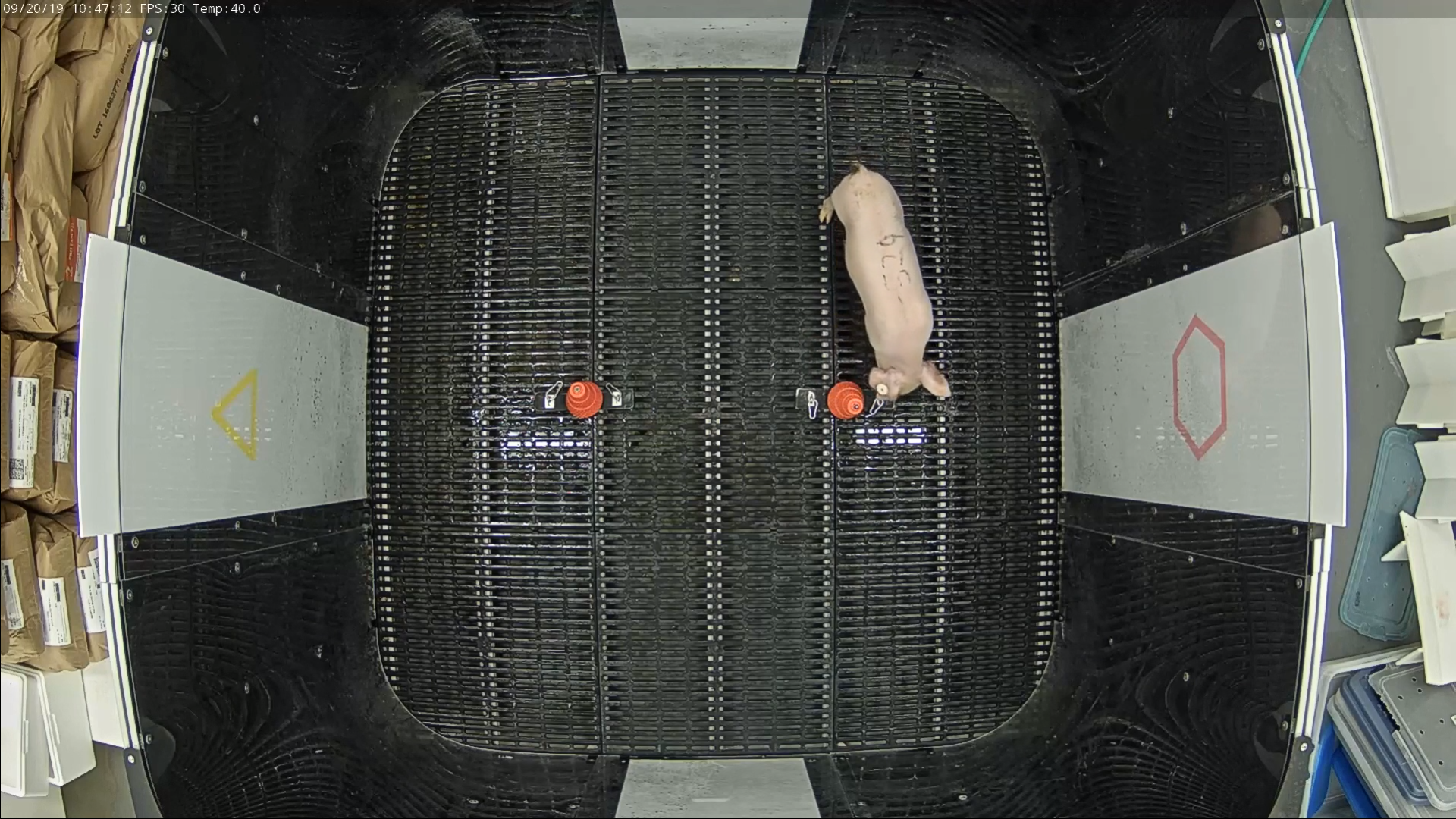}
\end{center}
   \caption{Novel Object Recognition arena with two objects and a single test subject}
   \label{fig:arena}
\end{figure}

A validated NOR task is used to assess recognition memory in pigs \cite{fleming2017young}. The performance on this task can be used to compare the development of recognition memory between control and test groups of pigs. In this task, a pig is freely allowed to explore a familiar environment that contains one familiar object and one novel object. The interactions with the two objects are of interest. Each NOR task is recorded for 5.5 minutes with a single piglet in the arena as depicted in Figure \ref{fig:arena}.  

Without any external learning models, the raw video files are analyzed manually to record the number of interactions with either object, the duration of each interaction, as well as the start and stop times of the trial. The following are the key metrics associated with every video, which are of interest in NOR analysis:
\begin{itemize}[noitemsep,topsep=0pt]
    \item \textbf{N}: Total number of investigations
    \item \textbf{CD}: Total time spent investigating
    \item \textbf{ME}: Average amount of time for any one investigation
    \item \textbf{LF}: Latency to the first investigation
    \item \textbf{LL}: Latency to the last investigation
    \item \textbf{RI} (Recognition Index): Amount of time spent investigating the novel object proportional to the total amount of time investigating both objects
\end{itemize}

These metrics allow for analysis of cognitive development and performance, with emphasis on RI as the main determinant. Even though there is a standard protocol for processing NOR videos, each human analyzer produces slightly different results, due to subjective interpretation of piglet actions. Moreover, each 5-minute video takes 30 minutes of manual effort to analyze and annotate. This serves as a motivation to develop an automated process that can standardize output and minimize human intervention.

\section{Action Recognition Methods}
\label{sec:models}

We explore three state-of-the-art action recognition methods: Long-term Recurrent Convolutional Networks (LRCN) \cite{Donahue}, 3D Convolutional Networks (C3D) \cite{tran2014learning} and Temporal Shift Module (TSM) \cite{lin2018tsm}.

\subsection{\textbf{LRCN}}

\begin{figure}[h]
    \centering
    \begin{subfigure}{.5\textwidth}
      \includegraphics[width=0.9\linewidth]{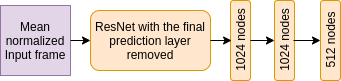}
    \caption{LRCN: CNN encoder}
    \label{fig:cnn-encoder}
    \end{subfigure}
    
    \begin{subfigure}{.5\textwidth}
      \includegraphics[width=0.9\linewidth]{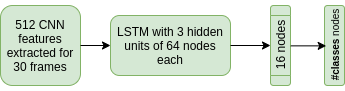}
    \caption{LRCN: RNN decoder}
    \label{fig:rnn-decoder}
    \end{subfigure}
  \caption{LRCN architecture}
\end{figure}

The broader idea for LRCN is to encode every frame using a Convolutional Neural Network (CNN) to capture spatial features in the frame and then pass features from consecutive frames through a Recurrent Neural Network (RNN) to capture the temporal variations in the frame. In our architecture, a pre-trained ResNet-18 network serves as a CNN feature extractor. Given an input image, a 512-dimensional feature vector is generated. The extracted features for 30 frames are stacked together and then passed through an LSTM to predict the action class for the stack of frames. 
Figure \ref{fig:cnn-encoder} depicts the architecture adopted for the CNN encoder and Figure \ref{fig:rnn-decoder} depicts the RNN decoder.

\subsection{\textbf{C3D}}

\begin{figure}[h]
\begin{center}
  \includegraphics[width=\linewidth]{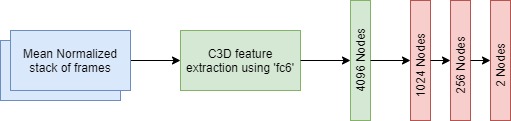}
\end{center}
  \caption{C3D architecture}
  \label{fig:c3d-arch}
\end{figure}

C3D is a state-of-the-art architecture used for learning spatio-temporal features by convolving over segmented volumes of videos. A C3D network pretrained on the Sports1M dataset is used to capture the spatio-temporal variation in a clip of 30 frames, and the output of the \texttt{fc6} layer is used to extract a single 4096-dimensional feature vector. A Binary Classifier is built which uses these feature vectors as input to predict the class label, as depicted in Figure \ref{fig:c3d-arch}.

\subsection{\textbf{TSM}}
Since 3D-CNN-based methods can achieve good performance but are computationally intensive, the RGB variant of TSM is explored for action recognition. The key idea in TSM is similar to LRCN, where features for each frame in a clip are extracted using a backbone CNN. But instead of training an RNN model, the features are pooled using a Temporal Shift Module to obtain a prediction for the clip. The model is trained using a TSM network with a backbone ResNet-18 network and takes in RGB sequences of frames as input.

\section{Pig Novelty Preference Behavior Dataset}

\subsection{Annotations}
The `Pig Novelty Preference Behavior' (PNPB) dataset is captured at the Pig Nutrition \& Cognition Lab at the University of Illinois, Urbana-Champaign. PNPB 
consists of a total of 20 videos, each of which is 5.5 minutes long in duration. The videos are captured at a resolution of 1024x1024 and at 30 frames per second. Each video contains only one pig, and all pigs are in the same environment. 

We provide two kinds of annotations for the following tasks:

\textbf{Action Recognition}: All the videos are annotated with time intervals for object investigations made by the pig for each object (identified as the left and the right object). The rest of the time intervals are identified as exploration. The annotations are made at the frame level, the highest granularity possible for pinpointing the object interactions. 

\textbf{Pig Keypoint Detection}: Nine videos are annotated with three keypoints on the pig. As depicted in Figure \ref{fig:hist-annotate}, the base of the tail, the tip of the nose, and the crown of the pig, as well as bounding box for the pig, are annotated for 668 frames. Since the focus of this paper is on action recognition, we do not use the keypoint annotations in our model.

\subsection{Post-processing}

\begin{figure}[h]
    \centering
    
    
    \includegraphics[width=\linewidth]{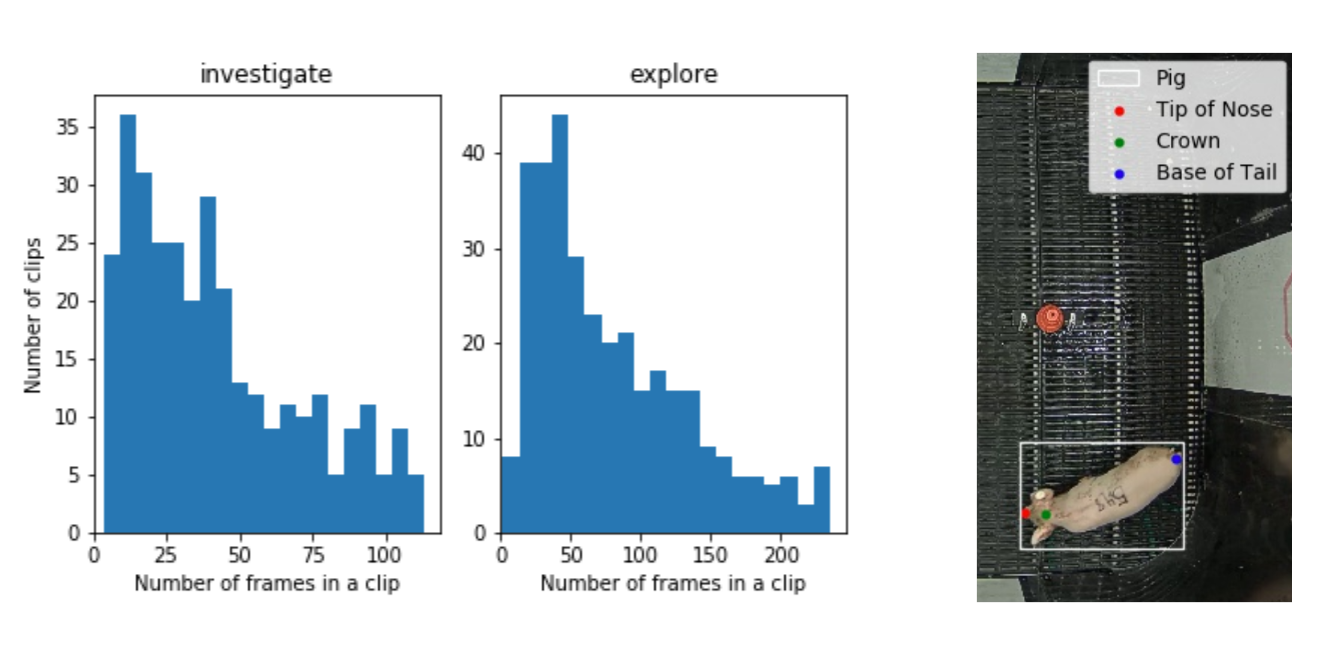}  
    \caption{(a) Action Histogram (b) Keypoint Annotations}
    \label{fig:hist-annotate}
\end{figure}

For the purpose of training action recognition models, the raw videos were pre-processed to extract clips of relevant actions using the time intervals specified in the annotations. Barring the outliers, a histogram of the number of clips for a given activity bucketed against the number of frames per clip shows that a pig can spend from a fraction of a second up to 3 seconds for a single object investigation, as depicted in Figure \ref{fig:hist-annotate}. To maintain consistency while training the deep learning models, all the clips which contained less than 60 frames are removed and all the bigger clips are fragmented into smaller clips having exactly 60 frames each. Thus, 1134 and 1109 clips are used for class `explore' and `investigate' respectively, and every clip contains 60 frames each for the initial experiments. 

\section{Evaluation}
\label{sec:results}
In this section, we evaluate the efficacy of the models described in the section \ref{sec:models}. To have a consistent notation, a single clip is defined as a contiguous set of frames (images), and a video is defined as a continuous set of clips.

\subsection{Training Procedure}

For the training procedure, a consistent training and validation split of 75\%-25\% is maintained. Hence, we use 1682 clips for training and 561 clips for validation.

\medskip
\noindent\textbf{LRCN}: Each frame is resized to 224x224 and normalized using the mean and standard deviation of the ImageNet \cite{imagenet} dataset. The cross-entropy loss is minimized for 30 epochs with a learning rate of 0.0001 using Adam optimizer.

\medskip
\noindent\textbf{C3D}: Each frame is resized to a resolution of 128x171 and normalized using the mean and standard deviation used for training the C3D model on the Sports1M dataset. C3D features for all the clips in the dataset are pickled and stored as a 4096-dimensional vector per clip. 

A binary classifier is trained by minimizing the binary cross-entropy loss for 20 epochs with a learning rate of 0.001 using Adam optimizer. Since the majority of the computation is done in the feature extraction step, the model is trained within few minutes on a CPU.

\medskip

\noindent\textbf{TSM}: Since ResNet-18 is used as the backbone network, similar preprocessing techniques are applied to the frames as seen in the case of LRCN. The cross-entropy loss is minimized for 35 epochs with an initial learning rate of 0.02 using an SGD optimizer with a momentum of 0.9 and a weight decay of 0.0005. The learning rate is adjusted to 0.002 and 0.0002 after 12 and 25 epochs, respectively. 

\subsection{Clip Level Metrics}

\begin{table}[h]
\centering
\begin{tabular}{@{}|c|c|c|c|c|c|@{}}
\hline
Model & Acc. & mAP & FPS & GMACs & ParamsM \\ 
\hline
LRCN  & 93\% & 89\% & 18.20 & 109.43 & 13.30 M \\
C3D   & \textbf{95\%} & 94\% & 110.84 & 17.5 & 65.67 M \\
TSM   & 93\% & \textbf{96\%} & \textbf{125.19} & \textbf{14.57} & \textbf{11.18 M} \\ 
\hline
\end{tabular}
\caption{Analytical and Computational Metrics}
\label{tab:clip-metrics}
\end{table}

Firstly, we evaluate how the models work on predicting individual clips of pigs. We consider clips of two seconds for this evaluation and Table~\ref{tab:clip-metrics} reports the accuracy values on the validation set of 561 clips which was constant across models. The Precision-Recall curve for the three models is computed by setting the class `investigate' as the positive label and the mean Average Precision (mAP) is reported. We observe that C3D and TSM are fairly comparable in terms of accuracy.

\begin{figure*}
    \centering
    \includesvg[width=\linewidth]{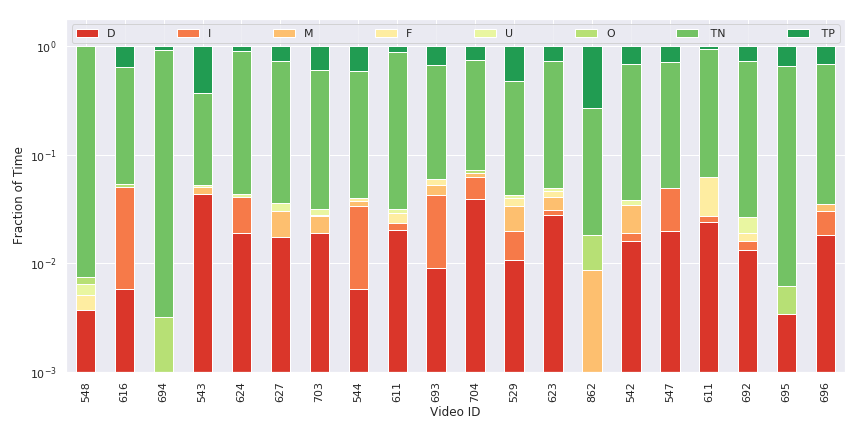}
    \caption{Continuous Action Recognition Metrics plotted on a log scale. The model yields less than 10\% errors, which are primarily the result of deletions. On average, a severe error is made 3\% of the times.}
    \label{fig:cont-metrics}
\end{figure*}

\subsection{Computational Performance}

We use an Intel(R) Xeon(R) CPU E5-2697 machine with 128GB RAM and 28 cores to measure the computational performance of different models on a CPU. The following metrics are compared for each model - frames per second achieved during inference (FPS), multiply-accumulate operations in billions (GMACs), and the number of model parameters in millions (ParamsM). The numbers for each model are reported in Table \ref{tab:clip-metrics}.

\subsection{Continuous Action Recognition Metrics}
As seen in the previous subsection, TSM has a computational advantage over other models while maintaining comparable accuracy. Hence, we conduct further experiments by using a TSM model trained on a clip size of 1 second.

Since the final NOR metrics described in Section \ref{sec:nor} depend on the intervals of investigation for the whole video, it is imperative to evaluate continuous action recognition metrics to see how models perform at video-level instead of clip-level. To that end, we report the performance metrics introduced in \cite{minnen2006performance}.

Since the model is trained using a clip size of 1 second, predictions are made at the granularity of seconds instead of frames. Considering the activity class `investigate' as the positive class, the predictions for each video in PNPB are split into mutually exclusive chunks of frames. True Positives (TP) and True Negatives (TN) are those frames where the predictions match the ground truth. Overfill (O) and Underfill (U) errors occur when an activity is correctly identified, but the boundaries of the predictions do not match that of ground truth exactly. Fragmentation (F) errors occur when a single ground truth has multiple fragmented predictions. On the contrary, Merges (M) happen when multiple ground truth intervals are combined into a single prediction. Insertions (I) and Deletions (D) are more severe errors where an action interval is falsely inserted or deleted respectively.

Figure \ref{fig:cont-metrics} depicts the metrics plotted on a log scale for all the videos in the PNPB dataset. One can see that the model is able to predict true positives and true negatives more than 90\% of the time. The chances of a severe error happening is around 3\%. As we see in the next subsection, the errors happening in this small fraction of time can cause the model to make wayward estimation of NOR-related metrics.

\subsection{NOR-related Metrics}
Clip-level and Video-level metrics are a measure of the performance of the model with respect to the ground truth annotations. But in order to fully automate the process of annotation, deep learning models must match the NOR metrics of a human annotator to a certain degree of confidence.

We post-process the clip-level predictions obtained by the model and the ground truth annotations to calculate NOR metrics described in Section \ref{sec:nor}. Since these metrics are continuous variables, one can think of the predictions obtained by the model as the output of a regression model, where the input is an NOR video. Hence, we report the R\textsuperscript{2} statistic along with the mean and standard deviation of error, i.e., $m_{predicted}-m_{groundtruth}$, where $m$ is an NOR metric.

\begin{table}[h]
\centering
\tabcolsep=0.13cm
\begin{tabular}{@{}|c|cccccc|@{}}
\hline
Statistic & N & CD  & ME & LF & LL & RI\\ \hline
Mean & 24.15 & 113.06 & 4.72 & 5.69 & 316.83 & 0.49 \\
\hline
R\textsuperscript{2}  & 0.58 & 0.99 & 0.77 & 0.42 & -0.26 & 0.86  \\
Mean Error & -4.0 & -1.74   & 0.90 & 2.49 & -6.72 & -0.03 \\
Std Error & 3.3 & 6.35 & 1.23 & 5.37 & 35.25 & 0.07 \\
\hline
\end{tabular}
\caption{Statistics related to predicted NOR metrics}
\label{tab:nor-metrics}
\end{table}

We report the mean value of all the metrics in Table~\ref{tab:nor-metrics}, in order to understand the scale of the metrics extracted from a single video. The R\textsuperscript{2} statistic for CD, RI and ME is relatively higher, as the model is able to successfully capture the investigation duration. It is notable that the model underestimates the number of investigations. This is due to the inability of the model to capture investigations that last less than a second. We observe a low R\textsuperscript{2} statistic for N, LF and LL, indicating that a naive application of action recognition is not sufficient to match the level of human annotators for this task. This is expected, as the model is optimized to classify behavior as the target and not regress NOR metrics on the input video.

\section{Conclusion}
Extracting valuable information about pigs with high precision from video datasets is a challenging problem. We introduce a fully annotated `Pig Novelty Preference Behavior' dataset, where the goal is to match the accuracy of a human annotator. Even though the application of state-of-the-art action recognition models achieves 95\% accuracy in classifying short clips of pig behavior, the models cannot be applied in an end-to-end pipeline to extract domain-specific metrics. In order to perform scoring on par with human annotators, it is imperative that models capture the rapid movements of pigs, which is inherent to the animal. Advanced deep learning models should be fine-tuned for the challenges discussed in this work.

{\small
\bibliographystyle{ieee_fullname}
\bibliography{egbib}
}

\end{document}